\documentclass[runningheads]{llncs}

\usepackage[mobile]{eccv}

\usepackage{eccvabbrv}

\usepackage{graphicx}
\usepackage{booktabs}
\usepackage{amsmath}
\usepackage{amssymb}
\usepackage{makecell}

\usepackage{listings}
\usepackage{multirow}
\usepackage{multicol}
\usepackage{wrapfig}

\usepackage{bigstrut}
\usepackage{rotating}
\usepackage{subcaption}

\usepackage{marvosym} 

\usepackage[accsupp]{axessibility}  

\usepackage{hyperref}

\usepackage{orcidlink}

\begin{document}

\title{IRSAM: Advancing Segment Anything Model for Infrared Small Target Detection} 

\titlerunning{IRSAM}

\author{Mingjin Zhang\inst{1}\orcidlink{0000-0002-1473-9784} \and
Yuchun Wang\inst{1}\Letter\and
Jie Guo\inst{1}\Letter\orcidlink{0000-0002-6223-5492} \and
Yunsong Li\inst{1} \and
Xinbo Gao\inst{1, 2}\orcidlink{0000-0003-1443-0776} \and
Jing Zhang\inst{3}\orcidlink{0000-0001-6595-7661}}

\authorrunning{M Zhang and Y Wang et al.}

\institute{Xidian University, Xi'an, China \\
\email{\{mjinzhang, 22011210918, jguo, ysli\}@xidian.edu.cn}
\and
Chongqing University of Posts and Telecommunications, Chongqing, China
\and
The University of Sydney, Sydney, Australia}

\maketitle

\begin{abstract}
  The recent Segment Anything Model (SAM) is a significant advancement in natural image segmentation, exhibiting potent zero-shot performance suitable for various downstream image segmentation tasks. However, directly utilizing the pretrained SAM for Infrared Small Target Detection (IRSTD) task falls short in achieving satisfying performance due to a notable domain gap between natural and infrared images. Unlike a visible light camera, a thermal imager reveals an object's temperature distribution by capturing infrared radiation. Small targets often show a subtle temperature transition at the object's boundaries. To address this issue, we propose the IRSAM model for IRSTD, which improves SAM's encoder-decoder architecture to learn better feature representation of infrared small objects. Specifically, we design a Perona-Malik diffusion (PMD)-based block and incorporate it into multiple levels of SAM's encoder to help it capture essential structural features while suppressing noise.
  Additionally, we devise a Granularity-Aware Decoder (GAD) to fuse the multi-granularity feature from the encoder to capture structural information that may be lost in long-distance modeling. Extensive experiments on the public datasets, including NUAA-SIRST, NUDT-SIRST, and IRSTD-1K, validate the design choice of IRSAM and its significant superiority over representative state-of-the-art methods. The source code are available at: \href{https://github.com/IPIC-Lab/IRSAM}{github.com/IPIC-Lab/IRSAM}.
  \keywords{Infrared small target detection \and Segment anything model \and Perona-Malik diffusion equation \and Granularity-Aware Decoder}
\end{abstract}

\section{Introduction}
\label{sec:intro}

Infrared small target detection (IRSTD) plays a crucial role in various real-world applications, including traffic management and maritime rescue \cite{deng2016small, teutsch2010classification, zhang2020empowering}. Infrared imaging uniquely captures thermal radiation, less affected by atmospheric scattering than visible light. Thus, under challenging visible light conditions like fog or rain, infrared images provide richer target information, making them more suitable for detecting obscured or indistinguishable targets, especially small ones. Consequently, IRSTD has been a prominent focus in computer vision for decades \cite{ISnet, ying2023mapping}.

The traditional IRSTD methods can be divided into three subgroups: filter-\begin{wrapfigure}{r}{0.65\textwidth}
 \vspace{-6mm}
 \includegraphics[width=0.9\linewidth]{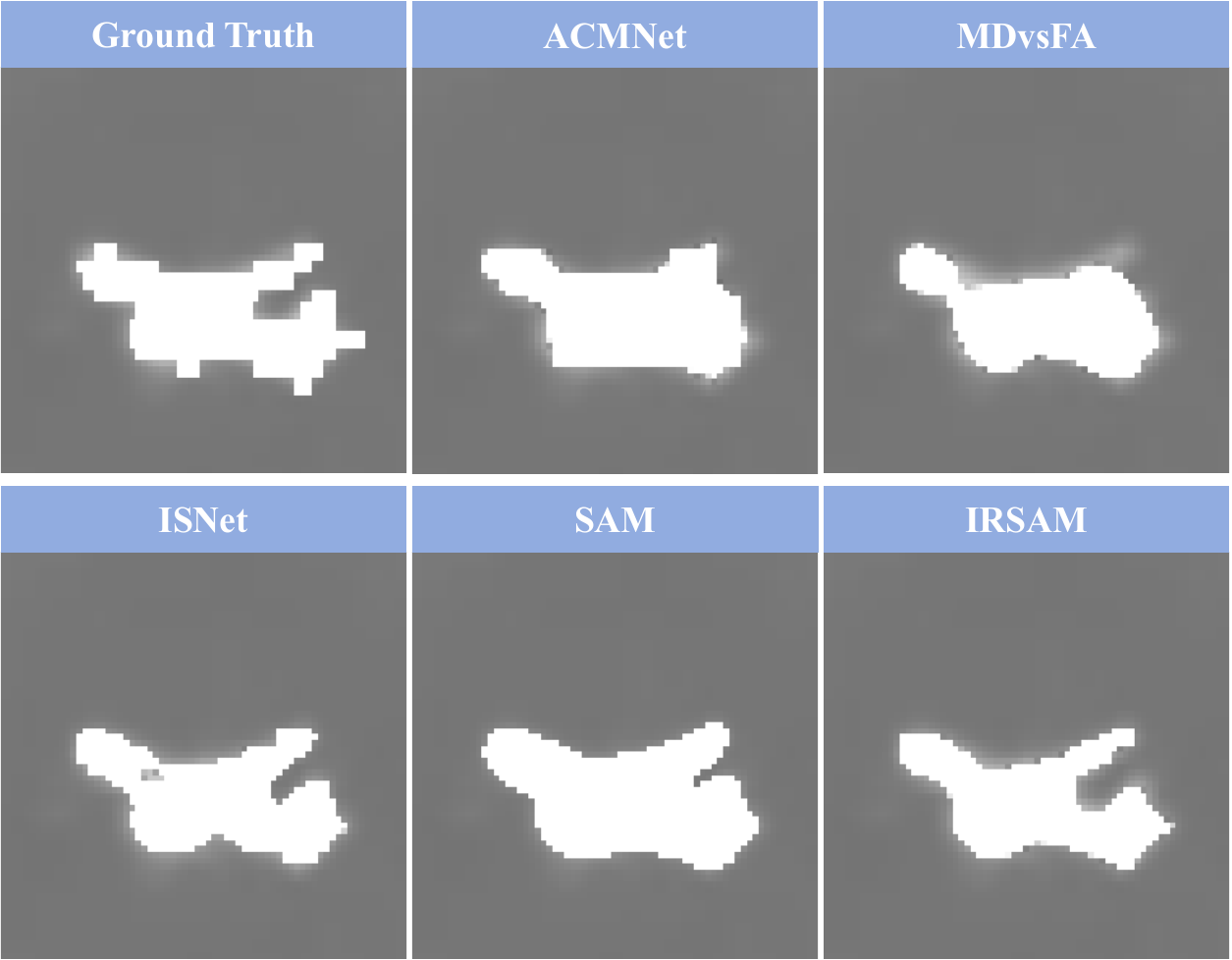}
 \caption{\textbf{Visual Comparison.} Segmentation results of different methods on complex structured targets. 
}
\label{fig_exhibition}
\vspace{-6mm}
\end{wrapfigure}
based, human visual system (HVS)-based, and low-rank represen-tation-based methods \cite{zhang2022exploring}. However, these methods are effective only in high contrast or  simple background scenarios, struggling in more challenging conditions due to heavy reliance on hyper-parameter tuning and hand-crafted features with limited representation ability. Recent advancements in deep learning and the availability of IRSTD public datasets offer a new solution. Approaches using deep learning for IRSTD include generative adversarial networks (GANs) \cite{zhang2019neural, zhang2019deep, gan, MDvsFA}, U-Net-based networks \cite{ACMNet, ALCNet, ISnet, UIUNet} and transformer-based methods \cite{zhang2022rkformer}.

Although these methods hold promise, their efficacy heavily relies on the architecture's specific design and the scale of training data. Obtaining such data is inherently more challenging compared to natural image datasets. With extensive research on deep models in the natural image domain and the proven effectiveness of transfer learning~\cite{transfer_1, transfer_2} in mitigating generalization issues with limited training data for downstream tasks, a crucial question arises: \textit{Can a well-designed model pre-trained on a large-scale natural image dataset effectively kickstart the IRSTD task?}

Segment Anything Model (SAM) \cite{SAM} has recently sparked a surge of interest in the field of computer vision image segmentation. Built upon the plain vision transformer-based encoder-decoder architecture and trained on the world's largest segmentation dataset, its powerful zero-shot segmentation capability has inspired many studies that apply SAM to various image segmentation tasks \cite{sam_medical_1, sam_medical_2, sam_medical_3, sam_hq}. While SAM shows promising results, they are not tailored for the IRSTD problem, facing challenges due to significant differences between infrared and natural images. (1) Infrared small target images, characterized by the small size of distant targets and substantial background noise and clutter, face challenges in distinguishing targets from the background when employing SAM directly due to a low signal-to-noise ratio (SNR). (2) Infrared imaging relies on the thermal radiation of objects, differing from optical imaging. The gradual radiation difference between the object and the background in infrared images results in a blurred target edge. This fuzzy edge makes SAM prone to segmenting the original gap into targets, especially when dealing with targets with complex structures, as demonstrated in \cref{fig_exhibition}. To address these issues, we aim to advance SAM for IRSTD through transfer learning, with a particular emphasis on refining its architecture to learn better feature representation for infrared small objects.

To this end, we propose the IRSAM model for IRSTD. It builds upon SAM with carefully designed blocks to improve its encoder and decoder, enhancing its ability to detect arbitrary infrared small objects in the context of background noise and clutter. Specifically, inspired by the Perona-Malik diffusion (PMD) equation used in image processing for image denoising and edge preservation, we design a Wavelet-based Perona-Malik Diffusion (WPMD) module by leveraging the wavelet transform to substitute the gradient term in the PMD equation. WPMD is incorporated into multiple levels of SAM's encoder to help it capture essential structural features while suppressing noise. Additionally, we devise a Granularity-Aware Decoder (GAD) to fuse the multi-granularity feature from the encoder via a two-way transformer to enhance the mask representation of objects in various sizes and shapes. To reduce the model size and computational complexity, we adopt the lightweight Mobile-SAM~\cite{mobile_sam} as the base model and integrated the WPMD module and GAD to build our IRSAM.

The contributions of this paper can be summarized as:
\begin{itemize}
    \item We introduce IRSAM by redesigning the general vision segmentation model SAM for the IRSTD task for the first time. IRSAM outperforms the vanilla SAM model and state-of-the-art (SOTA) methods on challenging benchmarks in terms of both objective metrics and subjective evaluation, demonstrating superior performance.
    \item We design a WPMD module to enhance the ability of SAM's encoder to preserve edge-related features while suppressing the noise in the infrared images, effectively addressing the low-SNR issue in the IRSTD task.
    \item We design an GAD to reconstruct the target structural feature lost in capturing long-distance dependence by fusing the multi-granularity feature from the encoder via a carefully designed edge token, effectively enhancing the mask representation of objects in various sizes and shapes.
\end{itemize}

\begin{figure*}[!t]
\centering
\includegraphics[width=1\linewidth]{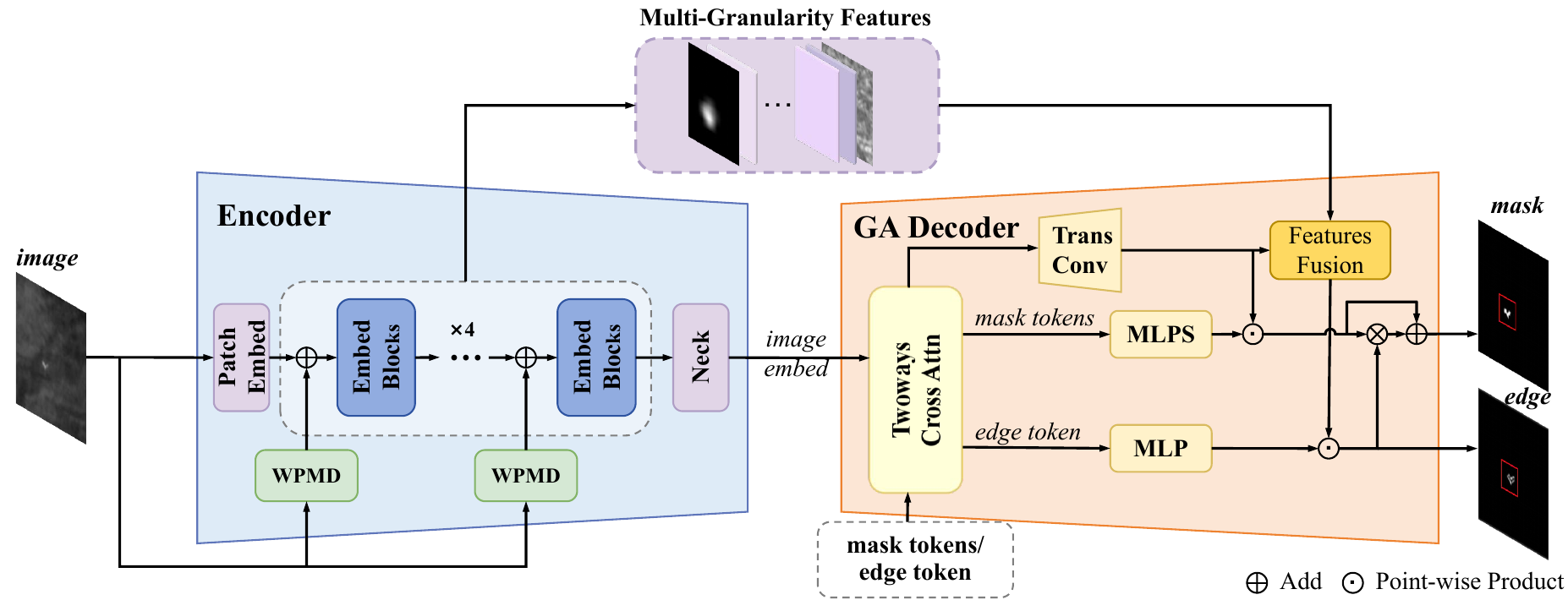}
\caption{\textbf{Overall Architecture of IRSAM}. Utilizing an encoder-decoder structure rooted in SAM, IRSAM incorporates two novel modules: WPMD and GAD, crafted specifically for the IRSTD task.
}
\label{fig_network}
\vspace{-2mm}
\end{figure*}

\section{Related Work}
\subsection{Infrared Small Target Detection}
The traditional IRSTD methods including Top-Hat \cite{Top_Hat}, Max-Median \cite{Max_Median}, WSLCM \cite{WSLCM}, TLLCM \cite{TLLCM}, IPI \cite{IPI}, NRAM \cite{NRAM}, RIPT \cite{RIPT}, PSTNN \cite{PSTNN}, and MSLSTIPT \cite{MSLSTIPT} based on hand-crafted features are tailored to specific scenarios, limiting their generalization to challenging situations. 
The deep learning-based methods have shown significant progress in IRSTD. For example, MDvsFA \cite{MDvsFA} applies conditional generative adversarial networks to address the IRSTD problem by achieving the trade-off between missed detection and false detection. Based on the U-Net structure, ACMNet \cite{ACMNet} introduces a novel feature fusion approach employing asymmetric context modulation, while UIUNet \cite{UIUNet} integrates multiple nested U-Net networks. In addition, Zhang \textit{et al.} \cite{zhang2022rkformer} present a transformer-based method for IRSTD called RKformer with random-connection attention. Considering the shape characteristics, Zhang \textit{et al.} \cite{ISnet} design a Taylor finite difference-inspired edge block, developing ISNet. To enhance SNR, Zhang \textit{et al.} \cite{dim2clear} devise a Dim2Clear from the perspective of image enhancement and super-resolution reconstruction. To explore a lightweight network architecture, Zhang \textit{et al.} \cite{irprunedet} make the first attempt to propose an IRPruneDet model tailored
for the IRSTD task through network pruning. Although these methods have shown promising results, they typically require large-scale labeled data for training. Obtaining such data is inherently more challenging compared to natural image datasets. Recently, Li \textit{et al.} \cite{li2023monte} and Ying \textit{et al.} \cite{ying2023mapping} recently introduce a cost-effective weakly supervised approach for IRSTD, leveraging single-point supervision to reduce annotation expenses. In this paper, we tackle this challenge from a different perspective. Given the extensive research on deep models in the natural image domain and the proven effectiveness of transfer learning~\cite{transfer_1, transfer_2} in mitigating generalization issues with limited training data for downstream tasks, we propose harnessing a foundation segmentation model pre-trained on a large-scale natural image dataset for the IRSTD task.

\subsection{Segment Anything Model}
Transformer \cite{Transformer, Zhang_2023_ICCV} has found widespread application in computer vision tasks, yielding competitive results in various domains. SAM, a Transformer-based model tailored for semantic segmentation \cite{SAM}, allows users to segment objects with prompts in any image. Following that, some researchers fine-tune the structure of SAM to improve its performance on complex tasks such as shadow detection and medical image detection \cite{sam_medical_1, sam_medical_2, sam_medical_3, sam_adapter}. To reduce the computation complexity of SAM, lightweight variants like Mobile SAM \cite{mobile_sam} and Fast SAM \cite{fast_sam} have been proposed. Some attempts have been made to enhance the decoder of SAM for improved segmentation under challenging conditions \cite{sam_hq}. However, these methods still rely on prompts to guide the segmentation process, restricting their applicability and efficiency for the IRSTD task. 
Besides, the inherent domain gap between natural and infrared images hampers SAM's performance in IRSTD. To address this challenge, we introduce the IRSAM model, enhancing SAM's encoder-decoder architecture to capture more effective feature representations for small infrared objects while leveraging its pre-trained knowledge for segmentation.

\subsection{Diffusion Equation for Image Processing}

The nonlinear diffusion equation, addressing anisotropic diffusion, has been used to adapt the diffusion coefficient based on local image features like gradient or curvature \cite{10586900, 9764823} to preserve edge and texture information while eliminating noise. PMD equation \cite{perona_malik} introduces gradient-based diffusion coefficient functions for image denoising and edge preservation. Chen \textit{et al.} \cite{diffusion_medical} apply PMD to medical image denoising, enhancing contrast and SNR by selecting suitable diffusion coefficient functions and parameters. Guo \textit{et al.} \cite{adaptive_PM} devise an adaptive PMD using variable exponent function spaces. Zhang \textit{et al.} \cite{zhang2022sar} propose a PMD neural module implemented in SAR-optical neural networks to reduce speckle noise in SAR images. Nevertheless, directly integrating PMD into neural networks presents challenges, including boundary blurring and reduced robustness to noise interference. In this work, we develop a WPMD module by incorporating the high-frequency component from wavelet transform as the image differential in the PMD equation, aiming to preserve structural information while mitigating noise impact.

\section{Methodology}
\subsection{Overall Architecture}
\cref{fig_network} illustrates the overall architecture of the proposed IRSAM, which adopts an encoder-decoder structure. The encoder of IRSAM comprises a pre-trained ViT-Tiny backbone and WPMD blocks. ViT-Tiny aggregates the edge features extracted by WPMD at each layer. For the decoder, unlike the original SAM decoder architecture, we fuse the features at different granularities from the encoder and use the output token to interact with image features and generate the final high-quality target mask.

\subsection{Wavelet transform-based PMD Block}
Perona-Malik diffusion equation \cite{perona_malik} is mainly used in image processing. The characteristic of anisotropic diffusion enables it to promote diffusion (smoothing) in smoother regions while suppressing diffusion at the edges, thereby achieving the effect of improving image quality, enhancing image structure, and suppressing noise\cite{perona_malik_denoise, adaptive_PM}. Infrared images are often corrupted by noise and have blurred object boundaries, which pose great challenges for the transfer of SAM to the IRSTD task. Therefore, we propose to explore PMD to preserve essential structural information and eliminate noises simultaneously in this transfer process. Instead of using the convolutions operator, we propose to use the high-frequency component from wavelet transform as the image differential in the PMD equation. As a result, the output of WPMD would be a smoother version of the input, preserving essential structural information while eliminating noise.

Given a picture \textit{u}, its corresponding PMD equation is given by:
\begin{equation}
\label{Eq:1}
    \frac{\partial u}{\partial t} = div\left( {g\left( \left| {\nabla u} \right| \right)\nabla u} \right),
\end{equation}
where diffusion coefficient $g\left( \left| {\nabla u} \right| \right) = 1/(1 + \left| {\nabla u} \right|^{2}/k^{2})$. $k$ is a positive constant used to control the degree of diffusion. $t$ represents the step \cite{adaptive_PM}. From \cref{Eq:1}, it can be seen that when the gradient magnitude $\left| {\nabla u} \right|$ of the smooth region is small, the diffusion coefficient $g\left( \left| {\nabla u} \right| \right)$ is large. Consequently, the diffusion is strong and the noise is effectively removed. In the edge part, the gradient magnitude $\left| {\nabla u} \right|$ is large and the diffusion coefficient $g\left( \left| {\nabla u} \right| \right)$ is small. Thus, the diffusion is weak and the edge information is retained.

\cref{Eq:1} can be expressed in the following form:
\begin{equation}
\begin{aligned}
\label{Eq:2}
    \frac{\partial u}{\partial t} &= \frac{\partial}{\partial x}\left\{ {g\left( \sqrt{\left( \frac{\partial u}{\partial x} \right)^{2} + \left( \frac{\partial u}{\partial y} \right)^{2}} \right)\frac{\partial u}{\partial x}} \right\}\\ &+ \frac{\partial}{\partial y}\left\{ {g\left( \sqrt{\left( \frac{\partial u}{\partial x} \right)^{2} + \left( \frac{\partial u}{\partial y} \right)^{2}} \right)\frac{\partial u}{\partial y}} \right\}.
\end{aligned}
\
\end{equation}

On the other hand, the 2D wavelet transform of an image can be represented:
\begin{equation}
\label{Eq:3}
    u_w = F_w (u), \ w \in \{LL, LH, HL, HH\},
\end{equation}
where $F\left(\cdot\right)$ represents the filtering operation. $L$ and $H$ stand for low-pass and high-pass, respectively.
Building upon the concept of approximating a differential equation$\frac{\partial}{\partial x}$ or $\frac{\partial}{\partial y}$ with a wavelet frame $F_{LH}(\cdot)$ and $F_{HL}(\cdot)$ as discussed in  \cite{wavelet_and_pde} and setting the diffusion step size $\triangle t$ to one, we can transform the \cref{Eq:2} into a discrete format:
\begin{equation}
\label{Eq:4}
u_{k} - u_{k - 1} = F_{LH}(g(\sqrt{u_{LH}^2+u_{HL}^2}) \cdot u_{LH}) + F_{HL}(g(\sqrt{u_{LH}^2+u_{HL}^2}) \cdot u_{HL}).
\end{equation}

\begin{figure}[!t]
\centering
\includegraphics[width=0.8\linewidth]{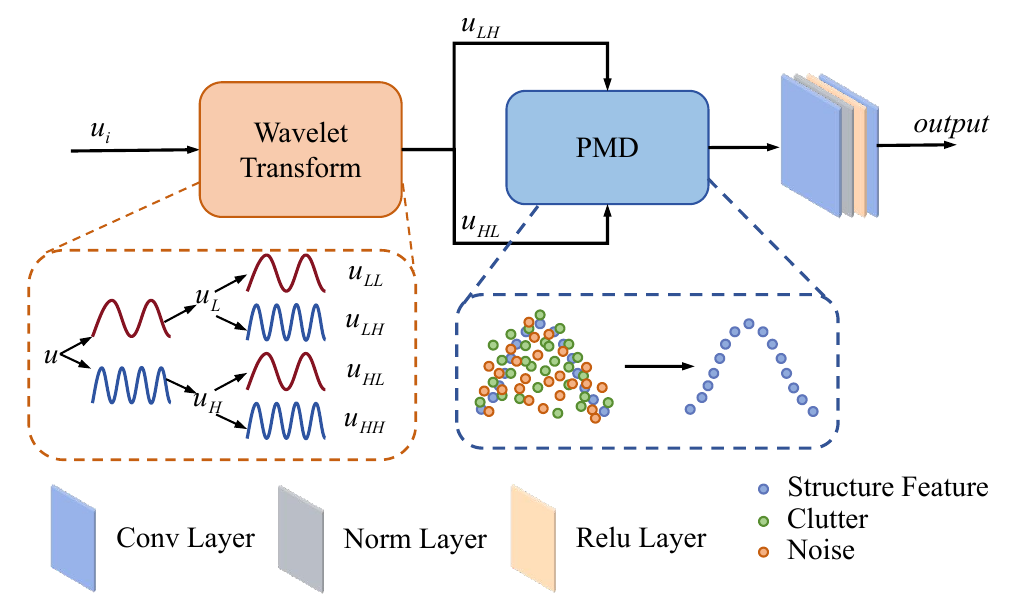}
\caption{Structure of the WPMD.}
\label{fig_WPMD}
\end{figure}

As shown in the \cref{fig_WPMD}, after the diffusion process, we use a convolutional layer to map the obtained structural features to the same dimension as the encoder features of each layer. By incorporating multiple WPMD modules into the SAM encoder at different layers, the SAM encoder enhances its ability to suppress noise while preserving the structural features of the infrared images, which effectively addresses the issue of low SNR and blurred target edge in the IRSTD task.

\subsection{Granularity-Aware Decoder}
Small infrared targets usually have limited visual features and are easily confounded with the background or similar targets. To improve the performance of infrared small target segmentation, it is necessary to consider both the global context information, which can help extract the overall semantics of the image and enhance the detection of small objects, as well as the local boundary information, which can help preserve the spatial details of small objects and improve the precision of segmentation boundaries. SAM adopts the ViT architecture, which excels at capturing long-term dependence and global information. The early layer of the ViT structure has been demonstrated to preserve more general image boundary details in previous works while the deep layer contains higher-level semantics \cite{ghiasi2022vision, sam_hq}. To improve the performance of SAM in the IRSTD task, we devise the Granularity-Aware Decoder to fuse the multi-granularity features By feeding global semantic context and local fine-grained features to the decoder, GAD enjoys a richer multi-view knowledge, as shown in \cref{fig_network}. 

First, we perform two-way cross attention on the image embeddings $X\in R^{64 \times 64 \times 256}$ from the encoder and output tokens $T_{tokens}$ consisting of mask tokens $T_{mask}\in R^{4 \times 256}$ and a newly designed edge token $T_{edge}\in R^{1 \times 256}$, \ie, image-to-token cross attention and token-to-image cross attention:

\begin{equation}
\label{Eq:5}
    X_{coarse},{\widetilde T}_{tokens}\;=\;\text{CrossAttention}(X,\;T_{tokens}).
\end{equation}

The ${\widetilde T}_{tokens}$ after the cross-attention update has integrated global image context as well as the information of other tokens and the edge token $T_{edge}$ is learnable and randomly initialized in the decoder like the mask tokens $T_{mask}$. We then upsample $X_{coarse}\in R^{64 \times 64 \times 256}$ and fuse it with the multi-granularity feature $X_{multi}$ to get the refined image features:
\begin{equation}
\label{Eq:6}
   X_{fine}=\text{TConv}(X_{coarse}) + X_{multi}.
\end{equation}
Here, $\text{TConv}$ denotes transposed convolution which is used to match the size of $X_{coarse}$ and $X_{multi}$.
$X_{multi}$ is obtained as follows:
\begin{equation}
\label{Eq:7}
\begin{aligned}   X_{multi}&= \text{Conv}(\text{ReLU}(\text{Norm}(\text{Conv}(X_{shallow}))))\\
    & + \text{TConv}(\text{ReLU}(\text{Norm}(\text{TConv}(X_{deep})))),
\end{aligned}
\end{equation}
where $X_{shallow}\in R^{128 \times 128 \times 128}$ and $X_{deep}\in R^{64 \times 64 \times 256}$ represent the features of the shallow layer (\ie, the first layer) and deep layer (\ie, the fourth layer) of the encoder, respectively. $\text{Conv}$, $\text{Norm}$, and $\text{ReLU}$ stand for convolution, layer normalization, and the ReLU activation function, respectively.

Finally, we use MLPs to generate dynamic convolution kernels from the updated tokens ${\widetilde T}_{tokens}\;$ in \cref{Eq:5} and apply them to $X_{fine}\in R^{256 \times 256 \times 32}$ and upsampled $X_{coarse}\in R^{256 \times 256 \times 32}$ in \cref{Eq:6} via Hadamard product to get the high-quality edge prediction and mask prediction. This process can be expressed:
\begin{equation}
    \label{Eq:8}
    X_{edge}=\text{MLP} \left({\widetilde T}_{edge}\;\right)\odot X_{fine},
\end{equation}
\begin{equation}
    X_{mask}=\text{MLP} \left({\widetilde T}_{mask}\;\right)\odot X_{coarse},
\end{equation}
where $\odot$ stands for dot product.
Finally, the obtained edge is used to improve the shape and size of the mask. Through the above process, the proposed GAD achieves the fusion of multi-granularity features in the decoder, integrates global context and local boundary information into the tokens, and finally obtains high-quality infrared small target masks.

\subsection{Loss Functions}
\textbf{Dice Loss:} Dice loss is a commonly used method to evaluate the difference between the predicted mask and the ground truth, which is defined as follows:
\begin{equation}
    L_{Dice}=1-2\times\frac{\left|Y_{Label}+Y_{Pred}\right|}{\left|Y_{Label}\right|+\left|Y_{Pred}\right|},
\end{equation}
where $\left| {Y_{Label} \cap Y_{Pred}} \right|$ is the intersection of the ground truth $Y_{Label}$ and the prediction $Y_{Pred}$. $\left|\cdot\right|$ is the number of target pixels in the mask.

\textbf{BCE Loss:} Binary Cross Entropy (BCE) Loss is a common loss function for binary classification tasks. It is used to measure the difference between the edge prediction $Y_{PredEdge}$ and the ground truth edge $Y_{Edge}$, which is defined as:
\begin{equation}
    L_{BCE} = - Y_{Edge}{{\times \log}{\left( Y_{PredEdge} \right)}} -\left( {1 - Y_{Label}} \right) \times {\log\left( {1 - Y_{PredEdge}} \right)}.
\end{equation}
The final loss consisting of the Dice loss $L_{Dice}$ and BCE loss $L_{BCE}$ is used to supervise the training of IRSAM:
\begin{equation}
    L = L_{Dice} + \lambda L_{BCE},
\end{equation}
where $\lambda$ is a hyper-parameter to balance the two losses and set to 10 empirically.

\begin{table*}[!t]
	\centering
	\caption{Comparison with representative methods on IRSTD-1k, NUDT-SIRST and NUAA-SIRST in $Params(M)$, $FPS(/s)$, $IoU(\%)$, $nIoU(\%)$, ${P_{d}}(\%)$, ${F_{a}}({10}^{-6})$.}
	\setlength{\tabcolsep}{0.5mm}
        \resizebox{\textwidth}{!}{
        \begin{tabular}{c||c|c||c|c|c|c||c|c|c|c||c|c|c|c}
		\hline
		  \multirow{2}{*}{Method} & \multirow{2}{*}{$Params$$\downarrow$} & \multirow{2}{*}{$FPS$$\uparrow$}& \multicolumn{4}{c||}{IRSTD-1k} & \multicolumn{4}{c||}{NUDT-SIRST} & \multicolumn{4}{c}{NUAA-SIRST}\\
		  \cline{4-15} &  &  & $IoU$$\uparrow$ & $nIoU$$\uparrow$ & ${P_{d}}$$\uparrow$ & ${F_{a}}$$\downarrow$& $IoU$$\uparrow$ & $nIoU$$\uparrow$ & ${P_{d}}$$\uparrow$ & ${F_{a}}$$\downarrow$ & $IoU$$\uparrow$ & $nIoU$$\uparrow$ & ${P_{d}}$$\uparrow$ & ${F_{a}}$$\downarrow$\\
		  \hline
		  Top-Hat \cite{Top_Hat} & - & - & 10.06 & 7.438 & 75.11 & 1432 & 22.40 & 37.56 & 89.90 & 174.1 & 1.508 & 3.084 & 79.74 & 16456 \\
		  Max-Median \cite{Max_Median} & - & - & 6.998 & 3.051 & 65.21 & 59.73 & 12.75 & 17.47 & 80.13 & 60.11 & 6.022 & 25.35 & 84.34 & 774.3 \\
		  WSLCM \cite{WSLCM} & - & - & 3.452 & 0.678 & 72.44 & 6619 & 1.809 & 7.258 & 75.89 & 595.3 & 6.393 & 28.31 & 88.74 & 4462 \\
		  TLLCM \cite{TLLCM} & - & - & 3.311 & 0.784 & 77.39 & 6738 & 1.683 & 6.977 & 75.56 & 1131 & 4.240 & 12.09 & 88.37 & 6243 \\
		  IPI \cite{IPI}  & - & - & 27.92 & 20.46 & 81.37 & 16.18 & 30.93 & 35.99 & 81.98 & 17.99 & 1.09 & 50.23 & 87.05 & 30467 \\
		  NRAM \cite{NRAM} & - & - & 15.25 & 9.899 & 70.68 & 16.93 &  6.93 & 6.19 & 56.40 & 19.27 & 13.54 & 18.95 & 60.04 & 25.23 \\
		  RIPT \cite{RIPT} & - & - & 14.11 & 8.093 & 77.55 & 28.31 & 29.67 & 37.57 & 91.65 & 65.30 & 16.79 & 20.65 & 69.76 & 59.33 \\
		  PSTNN \cite{PSTNN} & - & - & 24.57 & 17.93 & 71.99 & 35.26 & 27.86 & 39.31 & 74.70 & 94.31 & 30.30 & 33.67 & 72.80 & 48.99\\
		  MSLSTIPT \cite{MSLSTIPT} & - & - & 11.43 & 5.932 & 79.03 & 1524 & 8.34 & 7.97 & 47.40 & 881 & 1.080 & 0.814 & 0.052 & 8.183 \\
  	  \hline
            MDvsFA \cite{MDvsFA} & 3.92 & 139 & 49.50 & 47.41 & 82.11 & 80.33 & 75.14 & 73.85 & 90.47 & 25.34 & 60.30 & 58.26 & 89.35 & 56.35 \\
            ACMNet \cite{ACMNet} & 0.52 & 565 & 60.97 & 58.02 & 90.58 & 21.78 & 67.08 & 65.3 & 95.97 & 10.18 & 72.33 & 71.43 & 96.33 & 9.33 \\
            ALCNet \cite{ALCNet} & 0.54 & 534 & 62.05 & 59.58 & 90.58 & 21.78 & 81.40 & 80.71 & 96.51 & 9.26 & 74.31 & 73.12 & 97.34 & 20.21 \\
            Dim2Clear  \cite{dim2clear} & - & - & 66.34 & 64.27 & 93.75 & 20.93 & 81.37 & 80.96 & 96.23 & 9.17 & 77.29 & 75.24 & 99.10 & 6.72 \\
            UIUNet \cite{UIUNet} & 50.54 & 59 & 72.91 & 68.60 & 94.59 & 10.19 & 88.91 & 89.60 & 97.19 & 7.54 & 78.81 & 76.09 & 99.08 & 4.97 \\
            DNANet  \cite{DNANet} & 4.70 & 16 & 69.80 & 68.29 & 94.28 & 13.89 & 87.09 & 85.87 & 98.73 & 7.08 & 77.47 & 76.39 & 98.48 & 5.35 \\
            ISNet  \cite{ISnet} & 1.08 & 108 & 68.77 & 64.84 & 95.56 & 15.39 & 84.94 & 84.13 & 95.79 & 8.90 & 80.02 & 78.12 & 99.18 & 4.92 \\
		  \hline
		  \textbf{IRSAM (ours)}  & 12.33 & 103 & \textbf{73.69} & \textbf{68.97} & \textbf{96.92} & \textbf{7.55} & \textbf{92.59} & \textbf{93.29} & \textbf{98.87} & \textbf{6.94} & \textbf{80.78} & \textbf{78.39} & \textbf{99.53} & \textbf{3.95}\\
		\hline
	    \end{tabular}%
        }
	\label{tab1}%
\end{table*}%

\section{Experiments}
\subsection{Experiment Details}
\textbf{Datasets.} We perform experiments on three datasets, including NUAA-SIRST \cite{ACMNet}, IRSTD-1k \cite{ISnet}, and NUDT-SIRST \cite{DNANet}. They contain 427 and 1,000 real infrared images having one or more small targets, respectively, while NUDT-SIRST consists of 1,327 synthetic infrared images of small targets. All images in the datasets are resized to 512$\times$512. For each dataset, we use 50\% of the images as the train set, 30\% as the validation set, and 20\% as the test set, respectively.

\noindent\textbf{Evaluation Metrics.} We compare the proposed IRSAM against SOTA methods using pixel-level metrics such as \textit{Intersection over Union (IoU}) and \textit{Normalized Intersection over Union (nIoU)}\cite{ACMNet} and object-level metrics, including \textit{Probability of Detection ($P_{d}$)} and \textit{False-Alarm Rate ($F_{a}$)} for evaluation.

\noindent\textbf{Implementation Details}
We adopt the AdamW optimizer with a learning rate of 0.0001 and the Cosine Decay Learning Rate Scheduler to train our IRSAM. The model is trained for 500 epochs with a batch size of 4. The experiments are conducted on a single Nvidia GeForce 4090 GPU. For comparison, we choose CNN-based IRSTD methods: ISNet \cite{ISnet}, UIUNet \cite{UIUNet}, DNANet \cite{DNANet}, Dim2Clear \cite{dim2clear}, ALCNet \cite{ALCNet}, ACMNet \cite{ACMNet}, and MDvsFA \cite{MDvsFA}, and select traditional methods: Top-Hat \cite{Top_Hat}, Max-Median \cite{Max_Median}, WSLCM \cite{WSLCM}, TLLCM \cite{TLLCM}, IPI \cite{IPI}, NRAM \cite{NRAM}, RIPT \cite{RIPT}, PSTNN \cite{PSTNN}, and MSLSTIPT \cite{MSLSTIPT}.

\begin{figure*}[!t]
\centering
\includegraphics[width=1\linewidth]{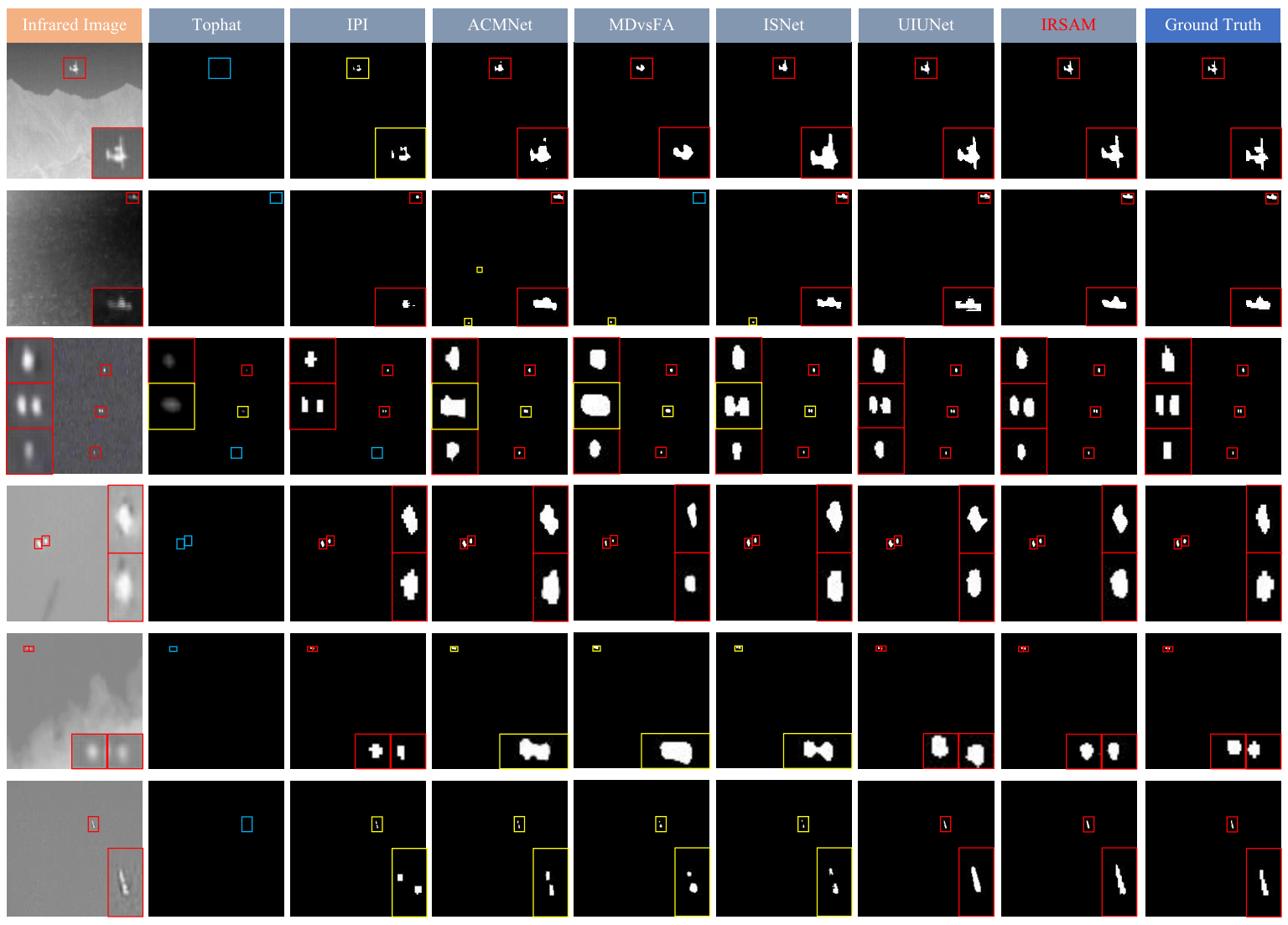}
\caption{Visualization results using different IRSTD methods. The closed views are shown at the border. In each prediction result, red, blue, and yellow boxes represent the correct detection, miss detection, and false detection, respectively.}
\label{fig_results}
\end{figure*}

\begin{figure*}[t!]
\centering
\includegraphics[width=1\linewidth]{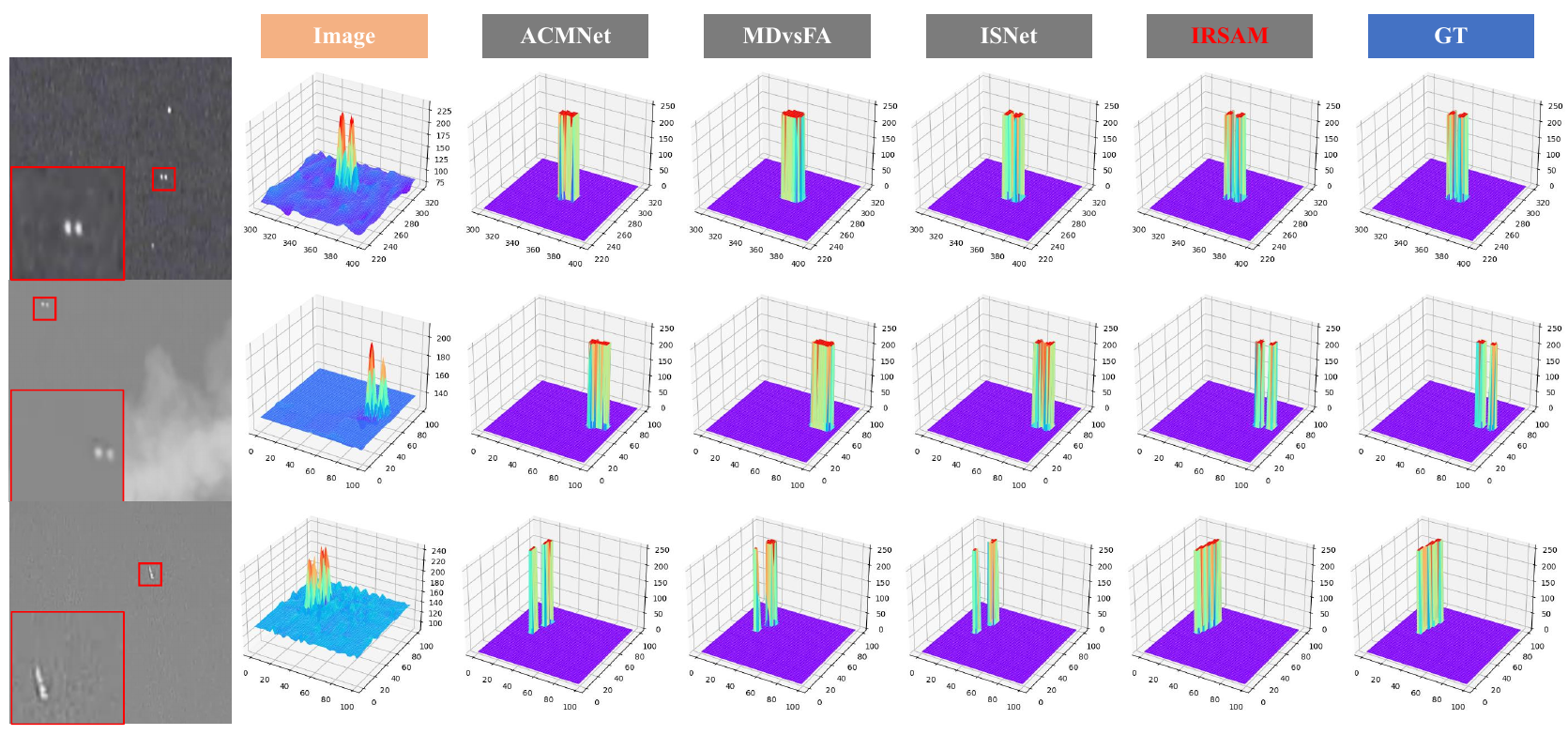}
\caption{3D views of the detection results obtained by different methods.}
\label{fig_3D}
\end{figure*}

\subsection{Quantitative Results}

As shown in \cref{tab1}, traditional hand-crafted features-based methods have limited capabilities in handling challenging scenarios, thereby yielding significantly worse performance compared to CNN-based methods. However, CNN-based methods exhibit limitations in detecting small targets, leading to inaccurate mask predictions with lower IoU and nIoU. Moreover, their efficiency in learning discriminative target representation is compromised in the presence of background noise, resulting in inefficient detection or missed detection. The proposed IRSAM outperforms SOTA methods in all evaluation metrics on NUAA-SIRST, IRSTD-1k, and NUDT-SIRST datasets. The results show that IRSAM can effectively extract the structural information of the target, which is attributed to the proposed WPMD and GAD that improve the learning ability of the vanilla SAM architecture for IRSTD as well as the strong generic segmentation ability of SAM.\begin{wrapfigure}{r}{0.6\textwidth}
    \centering
    \vspace{-6mm}
    \includegraphics[width=1\linewidth]{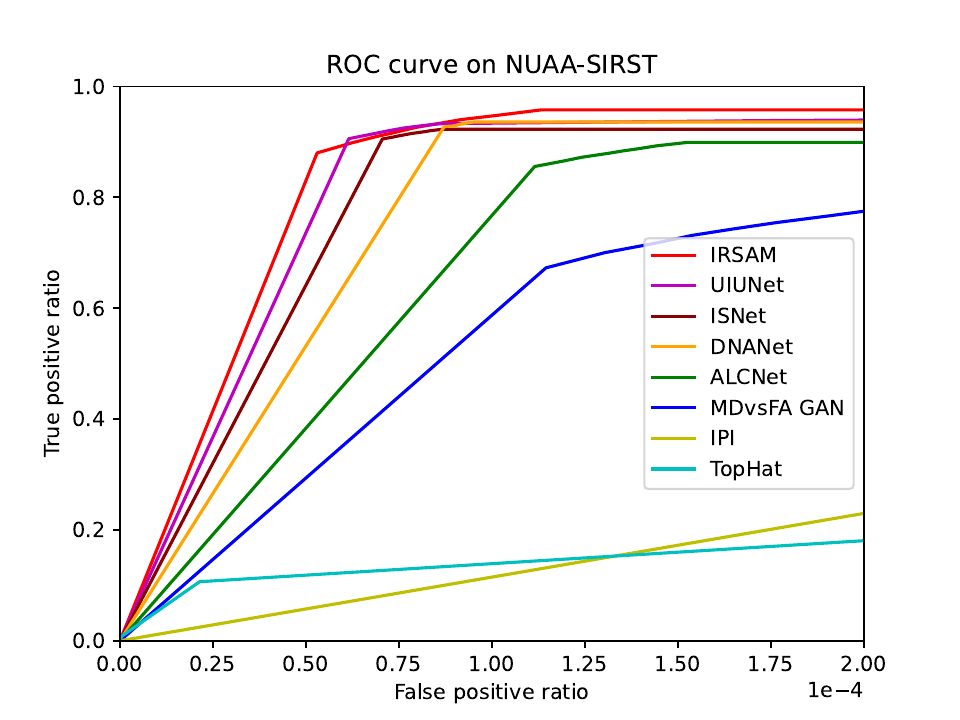}
    \caption{ROC curves of different methods.}
    \label{fig_roc_nuaa}
    \vspace{-6mm}
\end{wrapfigure}

We also plot the ROC curves of the results of different methods on NUAA-SIRST as shown in \cref{fig_roc_nuaa}. The results clearly demonstrate that our IRSAM outperforms the other methods by a substantial margin. The area under the ROC curve (AUC) for the proposed IRSAM is notably larger than that of both the traditional methods and the CNN-based methods.

As depicted in \cref{tab-SAM}, we compare our IRSAM with several SAM models that use different Transformer backbones with weights from MetaAI. The SAM models are adapted to the IRSTD task by fine-tuning the decoder or applying LoRA \cite{lora, adaptformer} on the encoder, while keeping the original encoder parameters frozen due to the prohibitive computation cost of fully fine-tuning. Notably, our IRSAM only employs a lightweight ViT-Tiny backbone, offering greater computational efficiency. IRSAM outperforms other SAM models across all metrics. In addition, our WPMD method demonstrates superior application of SAM to IRSTD tasks compared to other fine-tuning approaches, validating the effectiveness of the proposed modules in transferring a well-trained segmentation model from large-scale natural image datasets to IRSTD.

\begin{table}[t!]
    \caption{Comparison with fine-tuned SAM models on NUAA-SIRST in $IoU$, $nIoU$, ${P_{d}}$, ${F_{a}}$, $Flops (G)$, $Params (M)$.}
    \centering
     \setlength{\tabcolsep}{0.5mm}
    {\begin{tabular}{l||l|l|l|l|l|l}
    \hline
        \makecell[c]{Method} & \makecell[c]{$IoU$$\uparrow$} & \makecell[c]{$nIoU$$\uparrow$} & \makecell[c]{${P_{d}}$$\uparrow$} & \makecell[c]{${F_{a}}$$\downarrow$} & \makecell[c]{$Flops$$\downarrow$} & \makecell[c]{$Params$$\downarrow$}\\ \hline

        \makecell[c]{SAM(ViT-B) with finetuned decoder} & \makecell[c]{67.45} & \makecell[c]{64.41} & \makecell[c]{95.41} & \makecell[c]{51.05} & \makecell[c]{371} & \makecell[c]{90}\\ 
        \makecell[c]{SAM(ViT-L) with finetuned decoder} & \makecell[c]{69.55} & \makecell[c]{67.38} & \makecell[c]{98.17} & \makecell[c]{44.49} & \makecell[c]{1314} & \makecell[c]{308}\\ 
        \makecell[c]{SAM(ViT-H) with finetuned decoder} & \makecell[c]{70.96} & \makecell[c]{68.92} & \makecell[c]{97.25} & \makecell[c]{40.76} & \makecell[c]{2736} & \makecell[c]{635}\\ \hline

        \makecell[c]{SAM(ViT-B) with adaptformer \cite{adaptformer}} & \makecell[c]{67.90} & \makecell[c]{65.25} & \makecell[c]{95.41} & \makecell[c]{44.18} & \makecell[c]{372} & \makecell[c]{91}\\ 
        \makecell[c]{SAM(ViT-L) with adaptformer \cite{adaptformer}} & \makecell[c]{69.15} & \makecell[c]{65.59} & \makecell[c]{96.33} & \makecell[c]{40.50} & \makecell[c]{1315} & \makecell[c]{308}\\ 
        \makecell[c]{SAM(ViT-H) with adaptformer \cite{adaptformer}} & \makecell[c]{69.48} & \makecell[c]{66.31} & \makecell[c]{97.25} & \makecell[c]{26.8} & \makecell[c]{2738} & \makecell[c]{636}\\ \hline
        
        \makecell[c]{SAM(ViT-B) with WPMD} & \makecell[c]{74.15} & \makecell[c]{70.85} & \makecell[c]{97.25} & \makecell[c]{12.11} & \makecell[c]{611} & \makecell[c]{108}\\ 
        \makecell[c]{SAM(ViT-L) with WPMD} & \makecell[c]{75.04} & \makecell[c]{72.14} & \makecell[c]{98.17} & \makecell[c]{8.78} & \makecell[c]{1719} & \makecell[c]{338}\\ 
        \makecell[c]{SAM(ViT-H) with WPMD} & \makecell[c]{75.96} & \makecell[c]{73.48} & \makecell[c]{99.08} & \makecell[c]{5.65} & \makecell[c]{3341} & \makecell[c]{680}\\ \hline
        
        \makecell[c]{\textbf{IRSAM}} & \makecell[c]{\textbf{80.78}} & \makecell[c]{\textbf{78.39}} & \makecell[c]{\textbf{99.53}} & \makecell[c]{\textbf{3.95}} & \makecell[c]{\textbf{71.63}} & \makecell[c]{\textbf{12.33}}\\ \hline
    \end{tabular}}
    \label{tab-SAM}%
\end{table}

\subsection{Visual Results}
In \cref{fig_results}, we present some detection results obtained by IRSAM and other IRSTD methods. As shown in the first test image that contains an aircraft, most of the traditional methods and CNN-based methods suffer from false detection and fail to segment the target correctly. In contrast, the proposed IRSAM not only segments the target accurately but also outperforms other methods in the segmentation of the gap between the wing and the main body of the aircraft, indicating that the proposed method has a good ability to segment complex shape targets. In addition, from the 3rd and 5th test images, it can be observed that our method can successfully segment two adjacent objects, whereas other methods may fail. Besides, the result of the 6th figure shows that our method can make more realistic prediction for a long and narrow object.
We attribute the excellent performance of IRSAM in the above scenarios to the proposed WPMD block, which can effectively preserve the structural information of the target while suppressing the noise. Moreover, from all the given test images, it can be seen that the masks predicted by IRSAM are closer to the ground truth than other methods in both shape and completeness, validating the idea of introducing multi-granularity features to the decoder for obtaining higher-quality predictions. In addition, we show the 3D views of prediction results in \cref{fig_3D}. The proposed IRSAM performs well in segmenting multiple objects that are close to each other.

\subsection{Ablation Study}
\textbf{Impact of WPMD and GAD.}
As shown in \cref{tab-abalation1}, we conduct ablation studies to validate the effectiveness of WPMD and GAD on NUAA-dataset. For IRSAM without WPMD, we use the ViT-Tiny backbone for feature extraction in the encoder. Removing WPMD significantly reduces the $IoU$ and $nIoU$ scores, indicating the model's diminished ability to handle the target edge. $P_{d}$ remains stable while $F_{a}$ increases a lot. For IRSAM without GAD, we adopt the original decoder from SAM. Removing GAD reduces $IoU$ and $nIoU$ scores, validating the effectiveness of integrating multi-granularity features in the decoder for enhancing the segmentation performance. Furthermore, the Mobile-SAM baseline without both WPMD and GAD (first row) performs worse than the proposed IRSAM across all metrics. 

\begin{table}[t!]
    \caption{Ablation study of the WPMD and GAD in $IoU$, $nIoU$, ${P_{d}}$, ${F_{a}}$.}
    \centering
    \setlength{\tabcolsep}{5.5mm}
    \begin{tabular}{l||l|l|l|l}
    \hline
        \makecell[c]{Method} & \makecell[c]{$IoU$$\uparrow$} & \makecell[c]{$nIoU$$\uparrow$} & \makecell[c]{${P_{d}}$$\uparrow$} & \makecell[c]{${F_{a}}$$\downarrow$} \\ \hline
        \makecell[c]{w/o WPMD+GAD} & \makecell[c]{75.84} & \makecell[c]{73.29} & \makecell[c]{98.17} & \makecell[c]{8.21} \\ \hline      
        \makecell[c]{w/o GAD} & \makecell[c]{79.00} & \makecell[c]{76.29} & \makecell[c]{99.53} & \makecell[c]{2.13} \\ \hline
        \makecell[c]{w/o WPMD} & \makecell[c]{78.73} & \makecell[c]{76.39} & \makecell[c]{99.08} & \makecell[c]{11.18} \\ \hline
        \makecell[c]{IRSAM} & \makecell[c]{80.78} & \makecell[c]{78.39} & \makecell[c]{99.53} & \makecell[c]{3.95} \\ \hline
    \end{tabular}
    \label{tab-abalation1}%
\end{table}

\begin{figure}[!t]
\centering
\includegraphics[width=0.88\linewidth]{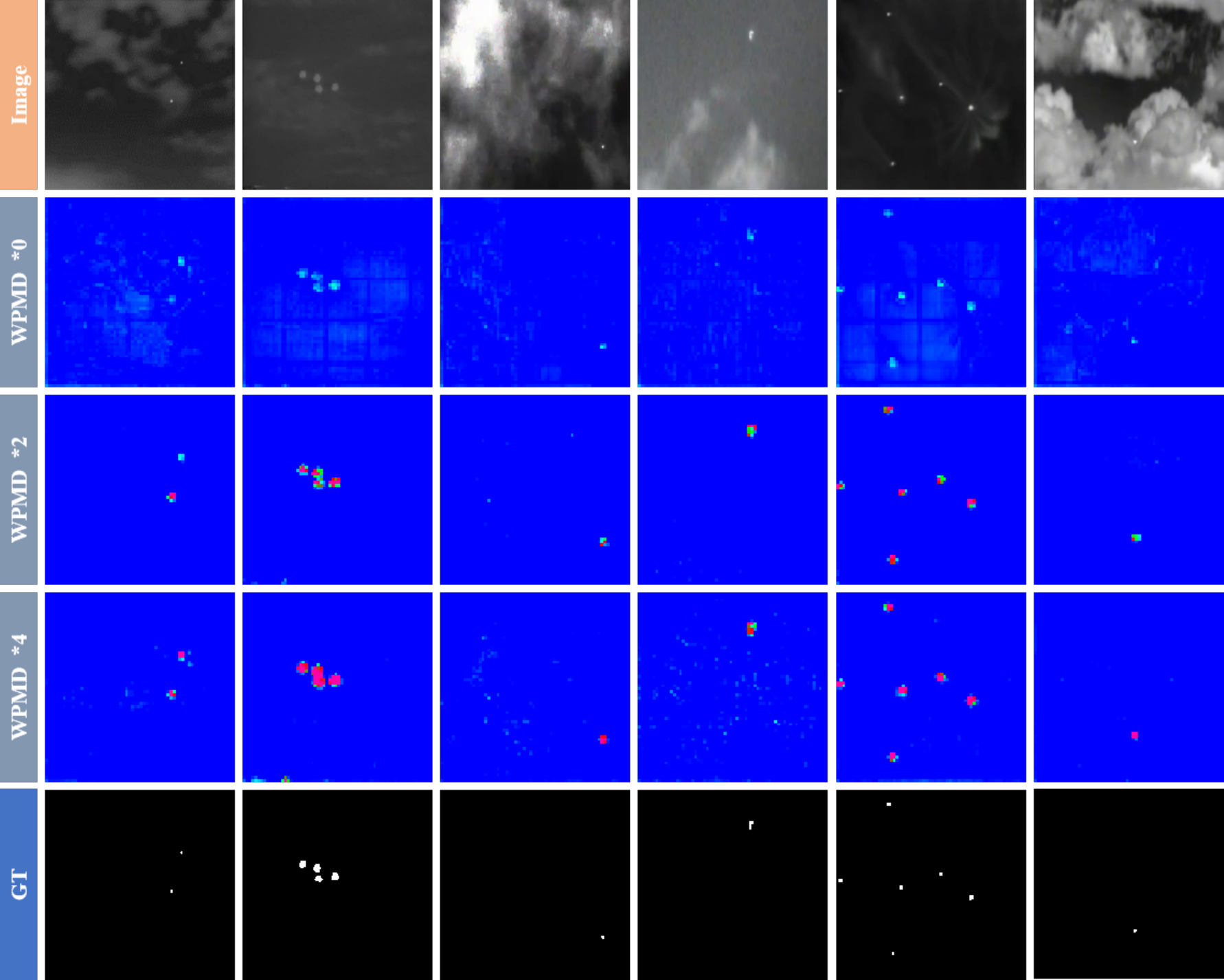}
\caption{Visualization of feature maps from encoders with different numbers of WPMD.}
\label{fig_feature}
\end{figure}

\begin{figure}[!t]
    \centering
    \includegraphics[width=0.86\linewidth]{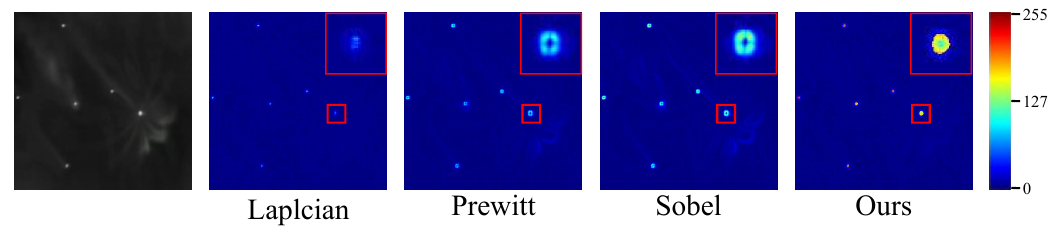}
    \caption{ Comparison of the edge maps utilizing various methods.}
    \label{fig:showedge}
\end{figure}

\begin{table}[t!]
    \caption{Ablation study on different number of WPMD blocks in $IoU$, $nIoU$, ${P_{d}}$, ${F_{a}}$.}
    \centering
    \setlength{\tabcolsep}{4.0mm}
    \begin{tabular}{l||l|l|l|l}
    \hline
        \makecell[c]{Number} & \makecell[c]{$IoU$$\uparrow$} & \makecell[c]{$nIoU$$\uparrow$} & \makecell[c]{${P_{d}}$$\uparrow$} & \makecell[c]{${F_{a}}$$\downarrow$} \\ \hline
        \makecell[c]{1} & \makecell[c]{76.17} & \makecell[c]{73.70} & \makecell[c]{98.17} & \makecell[c]{10.16} \\ \hline
        \makecell[c]{2} & \makecell[c]{77.69} & \makecell[c]{75.15} & \makecell[c]{99.08} & \makecell[c]{9.76} \\ \hline
        \makecell[c]{3} & \makecell[c]{78.25} & \makecell[c]{76.01} & \makecell[c]{99.08} & \makecell[c]{6.12} \\ \hline
        \makecell[c]{4} & \makecell[c]{79.00} & \makecell[c]{76.28} & \makecell[c]{99.53} & \makecell[c]{2.13} \\ \hline
    \end{tabular}
    \label{tab-abalation3}%
\end{table}

\begin{table}[t!]
    \caption{Ablation study on the design choice of GAD in $IoU$, $nIoU$, ${P_{d}}$, ${F_{a}}$.}
    \centering
    \fontsize{9}{11}\selectfont
    \setlength{\tabcolsep}{2mm}
    \begin{tabular}{l||l|l|l|l}
    \hline
        \makecell[c]{Type} & \makecell[c]{$IoU$$\uparrow$} & \makecell[c]{$nIoU$$\uparrow$} & \makecell[c]{${P_{d}}$$\uparrow$} & \makecell[c]{${F_{a}}$$\downarrow$} \\ \hline
        \makecell[c]{baseline} & \makecell[c]{78.64} & \makecell[c]{76.34} & \makecell[c]{99.08} & \makecell[c]{5.50} \\ \hline
        \makecell[c]{+ shallow feature} & \makecell[c]{79.22} & \makecell[c]{76.58} & \makecell[c]{99.08} & \makecell[c]{4.52} \\ \hline
        \makecell[c]{+ deep feature} & \makecell[c]{79.58} & \makecell[c]{77.16} & \makecell[c]{99.08} & \makecell[c]{4.17} \\ \hline
        \makecell[c]{+ shallow\&deep features} & \makecell[c]{80.78} & \makecell[c]{78.39} & \makecell[c]{99.53} & \makecell[c]{3.95} \\ \hline
    \end{tabular}
    \label{tab:gad}%
    \vspace{-1 mm}
\end{table}

\noindent\textbf{Impact of the Number of WPMD Modules.}
We also conduct ablation experiments to investigate the effect of using different numbers of WPMD modules in the encoder. As can be seen from \cref{tab-abalation3}, when the number of WPMD feature blocks is 4, the performance is better than others. This conclusion can also be verified by investigating the feature maps obtained by different numbers of WPMD. As shown in \cref{fig_feature}, when the number of WPMD blocks is 4, the target features in the obtained feature maps are significantly stronger than others. In addition, as visualized in \cref{fig:showedge}, the use of WPMD effectively preserves internal details and eliminates noises, achievements not paralleled by the Laplacian or Sobel operators . This comparison further demonstrates the effectiveness of our method in preserving essential structural information.

\noindent\textbf{Impact of design choice of GAD}
We also conduct ablation experiments to investigate the effect of encoder layer information for GAD. As shown in \cref{tab:gad}, When GAD leverages both shallow and deep features simultaneously, it outperforms other options. This suggests that GAD benefits from a richer multiview understanding by providing both global semantic context and local fine-grained features, resulting in a significant performance boost.

\section{Conclusion}

This paper introduces IRSAM, a novel approach for the IRSTD task. Leveraging the generic segmentation capability of the SAM foundation model, which is trained on a large scale of natural images, IRSAM enhances performance for IRSTD through two specifically designed modules: WPMD and GAD. WPMD improves the encoder's edge feature extraction, while GAD integrates multi-granularity features in the decoder for enhanced shape representation. Experimental results on public datasets, including NUAA-SIRST, IRSTD-1k, and NUDT-SIRST, demonstrate IRSAM's superiority over state-of-the-art methods in both objective metrics and subjective evaluation.

\section*{Acknowledgement}
This work was supported in part by the National Natural Science Foundation of China under Grant 62272363, Grant 62036007, Grant U21A20514, Grant 62061047, and Grant 62076190; in part by the National Key Research and Development Program of China under Grant 2023YFA100860; in part by the Young Elite Scientists Sponsorship Program by China Association for Science and Technology (CAST) under Grant 2021QNRC001; in part by the Joint Laboratory for Innovation in Satellite-Borne Computers and Electronics Technology Open Fund 2023 under Grant 2024KFKT001-1; in part by the Key Industrial Innovation Chain Project in Industrial Domain under Grant 2022ZDLGY01-11, and in part by the Key Research and Development Project of Xi’an under Grant 23ZDCYJSGG0022-2023.

%
%

\end{document}